%% file: main.tex
\documentclass[10pt,twocolumn,letterpaper]{article}

\usepackage[pagenumbers]{cvpr} 


\definecolor{cvprblue}{rgb}{0.21,0.49,0.74}
\usepackage[pagebackref,breaklinks,colorlinks,allcolors=cvprblue]{hyperref}

\def\paperID{MTF-18} 
\def\confName{CVPR}
\def\confYear{2025}

\title{SAMJAM: Zero-Shot Video Scene Graph Generation for Egocentric Kitchen Videos}

\author{Joshua Li\textsuperscript{1}, Fernando Jose Pena Cantu\textsuperscript{1}, Emily Yu\textsuperscript{1}, Alexander Wong\textsuperscript{1}, Yuchen Cui\textsuperscript{2}, Yuhao Chen\textsuperscript{1}\thanks{Corresponding author}\\
    \textsuperscript{1} Vision and Image Processing Lab, University of Waterloo \\
    \textsuperscript{2} Robot Intelligence Lab, UCLA \\
    {\tt\small \{j234li, fjpenaca, e33yu, alexander.wong\}@uwaterloo.ca} \\
    {\tt\small yuchencui@cs.ucla.edu} \\
    {\tt\small yuhao.chen1@uwaterloo.ca}
}

\begin{document}
\maketitle
\input{sec/0_abstract}    
\input{sec/1_intro}
\input{sec/2_method}
\input{sec/3_experiments}
\input{sec/4_conclusion}

{
    \small
    \bibliographystyle{ieeenat_fullname}
    \bibliography{main}
}


\end{document}

%% file: sec/0_abstract.tex
\begin{abstract}

Video Scene Graph Generation (VidSGG) is an important topic in understanding dynamic kitchen environments. Current models for VidSGG require extensive training to produce scene graphs. Recently, Vision Language Models (VLM) and Vision Foundation Models (VFM) have demonstrated impressive zero-shot capabilities in a variety of tasks. However, VLMs like Gemini struggle with the dynamics for VidSGG, failing to maintain stable object identities across frames.
To overcome this limitation, we propose SAMJAM, a zero-shot pipeline that combines SAM2's temporal tracking with Gemini's semantic understanding. SAM2 also improves upon Gemini's object grounding by producing more accurate bounding boxes. In our method, we first prompt Gemini to generate a frame-level scene graph. Then, we employ a matching algorithm to map each object in the scene graph with a SAM2-generated or SAM2-propagated mask, producing a temporally-consistent scene graph in dynamic environments. Finally, we repeat this process again in each of the following frames. We empirically demonstrate that SAMJAM outperforms Gemini by 8.33\% in mean recall on the EPIC-KITCHENS and EPIC-KITCHENS-100 datasets.

\end{abstract}

%% file: sec/1_intro.tex
\section{Introduction}
\label{sec:intro}

\begin{figure}[t]
  \centering
   \includegraphics[width=1\linewidth]{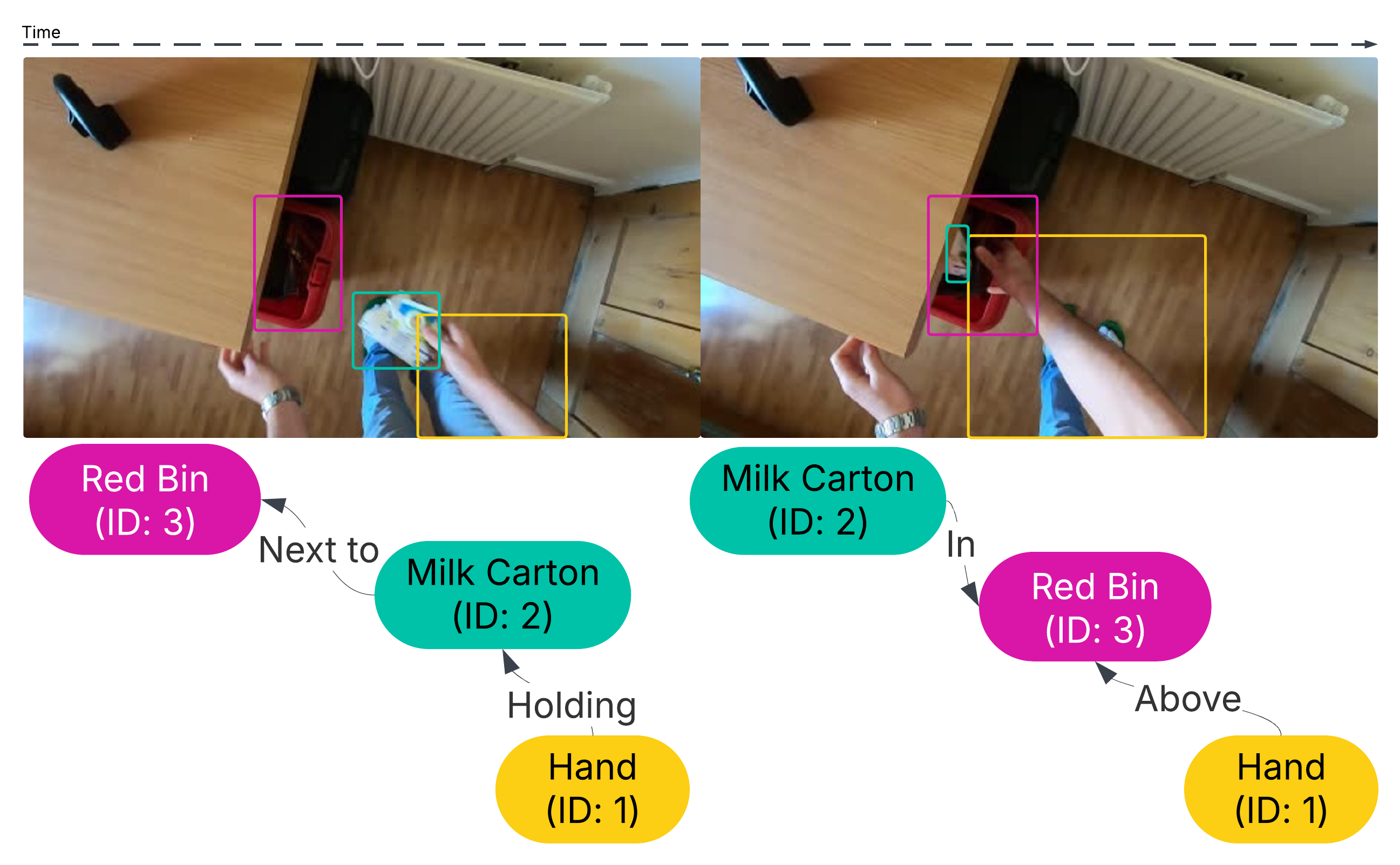}

   \caption{A video scene graph captures evolving relationships between objects in a dynamic environment. Video Scene Graph Generation (VidSGG) involves generating many frame-level scene graphs, each containing a set of objects and relationships. Objects that share a common ID across multiple frames are the same object temporally.}
   \label{fig:vidsgg}
\end{figure}

Understanding kitchen scenes from egocentric videos is an important challenge in computer vision as it opens the door to a wide range of applications, such as generating recipes with precise caloric and nutritional values directly from cooking footage \cite{skeleton_recipe}, providing robotic systems with the contextual awareness needed for culinary tasks \cite{Hori2023StyletransferBS}, and enabling more precise visual question answering for food-related queries. One way of representing a scene is via a \textbf{scene graph}. Scene graphs are structured representations where nodes denote objects and directed edges denote the relationships between them. Scene graphs offer high-level semantic information that bridges vision and language, supporting downstream tasks like image captioning, retrieval, and visual question answering \cite{Yang_Peng_Li_Guo, LI2024127052}.

Extending the scene graph concept to the temporal domain, \textbf{Video Scene Graph Generation (VidSGG)} seeks to parse a video into a dynamic sequence of scene graphs that captures objects and their evolving relationships over time (see Figure \ref{fig:vidsgg}). While static image-based scene graphs focus on snapshot-like relationships, VidSGG must account for object motion, appearance changes, and interaction dynamics \cite{Shang_Ren_Guo_Zhang_Chua_2017, Yang_Peng_Li_Guo, kim2025weakly}. 

Modern VidSGG methods often employ deep neural networks (e.g. CNNs and Transformers) to jointly track objects and infer relations while integrating temporal context \cite{yang2018graph, 10375886, Yang_Peng_Li_Guo, kim2025weakly}. More recently, Vision Language Models (VLMs) have shown impressive zero-shot capabilities \cite{Nagar_Jaiswal_Tan_2024, vlm_survery}, making them attractive candidates for open-vocabulary zero-shot VidSGG. While current VLMs like Gemini 2.0 Flash demonstrate good performance in long format video understanding -- as shown in the EgoSchema \cite{mangalam2023egoschema} and LongVideoBench \cite{wu2024longvideobench} benchmarks -- Gemini 2.0 Flash falls short when it comes to video scene graph generation. Specifically, Gemini has difficulty maintaining stable object identities across frames and producing precise bounding boxes at each frame. For example, given the same object in the video, Gemini could assign it an ID of 2 in frame 1 but an ID of 5 in frame 2. The bounding box of the object may also be too large or too small instead of being fitted around the object itself.

To address these challenges, we present SAMJAM, an innovative pipeline that integrates Segment Anything Model 2 (SAM2) \cite{ravi2025sam} with Gemini 2.0 Flash \cite{google2025gemini} to achieve temporally-consistent video scene graph generation. SAMJAM leverages SAM2’s robust segmentation and tracking capabilities to enhance both the spatial and temporal grounding of objects initially detected by Gemini. By merging Gemini’s zero-shot scene understanding with SAM2’s refined mask propagation, our method significantly improves VidSGG dynamics by maintaining stable object identities, ensuring precise object grounding, and continuously updating relationships throughout the video.

%% file: sec/2_method.tex
\section{Method}
\label{sec:method}

\begin{figure*}
  \centering
   \includegraphics[width=1.0\linewidth]{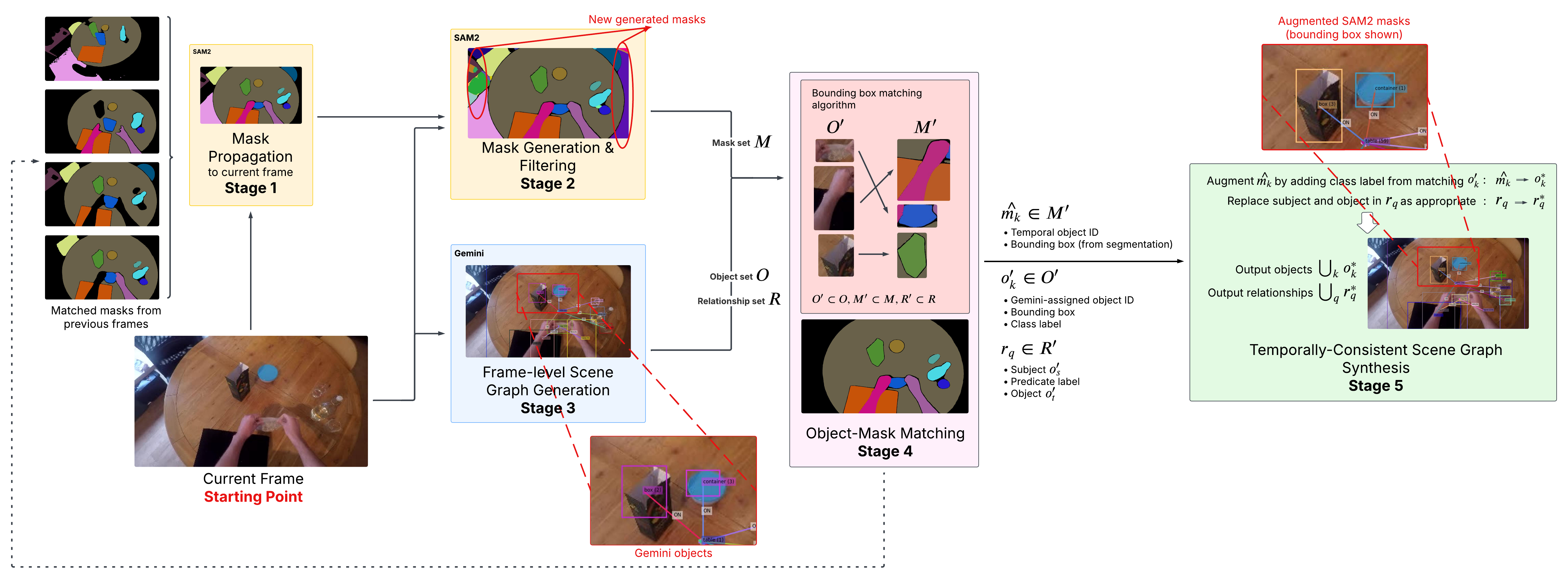}

   \caption{
   SAMJAM is a 5-stage pipeline at every frame. Given matched masks from earlier frames, SAM2 propagates masks to the current frame in stage 1. In stage 2, SAM2 generate a set of new masks and combines them with propagated masks, filtering out any overlap. In stage 3, Gemini independently produces a frame-level scene graph. We employ a matching algorithm in stage 4 that maps each Gemini object to a SAM2 mask, and finally synthesize a temporally-consistent scene graph in stage 5. To illustrate the transition from Gemini to SAM2, we also zooms in on two scene graphs produced along the pipeline. See Sec. \ref{first_frame} and Sec. \ref{following_frames} for details.}
   \label{fig:architecture}
\end{figure*}

\subsection{Problem Formulation}

We make slight adjustments to the problem formulation presented in papers \cite{Ji_2020_CVPR, nguyen2024cyclo}. Formally, VidSGG expects an input video sequence $v=\{f_1, f_2,...,f_n\}$, where each $f_i$ represents a single frame from the video. For each frame, the model is expected to generate a separate scene graph $G_i=(O, R)$. Each $o_k \in O$ represents an object with a class label and bounding box coordinates $o_k^{b}\in \mathbb{N}^4$, and each $r_q \in R$ represents a triplet $\langle o_s, p_q, o_t\rangle$ that denotes the relationship predicate $p_q$ between subject $o_s$ and object $o_t$. We further specify a set of global object IDs $I$ across all frames, such that each frame-level object $o_k$ is associated with an ID $o_k^I \in I$. This ID is used to indicate that objects on different frames are the same object temporally. Moreover, we expect such object IDs to be consistent in dynamic environments.

\subsection{Pipeline Overview}
For each frame, the SAMJAM pipeline is characterized by 5 stages: 1) mask propagation, 2) mask generation \& filtering, 3) frame-level scene graph generation, 4) object-mask matching, and 5) temporally-consistent scene graph synthesis. A visualization of the pipeline can be found in Figure \ref{fig:architecture}.

Given an input frame, Gemini produces a semantically-consistent scene graph that contains all relevant objects and relationships in the frame. However, without temporal context, Gemini-assigned object IDs are meaningless across different frames. Additionally, Gemini often fails to produce tight bounding boxes when grounding objects. SAM2 solves these issues by offering consistent video-level object tracking and precise frame-level object segmentation. Crucially, SAM2 dependably tracks objects even in dynamic environments. At a high level, we use SAM2 to propagate previous masks to the current frame, sample a new set of masks in the current frame, then filter out any overlap. Matching Gemini objects to the resulting masks produces a scene graph that takes advantage of both SAM2's refined bounding boxes and its temporally-consistent object IDs. This is our temporally-consistent scene graph.

Before arriving at SAMJAM, we considered many other approaches. For instance, an alternative solution begins with generating a set of SAM2 masks at each frame, then feeding the mask segmentations with the original image to Gemini. The hope was for Gemini to identify objects and relationships based on multi-image inputs that share a common spatial context. However, this approach yielded poor results, partly due to Gemini's inability to reason spatially over multiple images. Other approaches that rely solely on Gemini (or other VLMs) fail as well; see Sec. \ref{results} for a more in-depth discussion.

\subsection{Base Scene Graph Generation} \label{first_frame}
Consider the first frame $f_1$. Since there are no masks to propagate from earlier frames, we consider $f_1$ as the base case in our SAMJAM pipeline. Here, we outline a trimmed version of the 5-stage process (shown in Figure \ref{fig:architecture}) to build a foundation for mask-propagated VidSGG in later frames. 

To start, we skip mask propagation in stage 1 and move directly to mask generation in stage 2. We use SAM2's automatic mask generator class to generate a large set of (possibly overlapping) masks $M$. This is done via grid-based sampling over the image. For each SAM2-generated mask, we assign a temporal object ID (tID) that is unique to the mask both in the frame and across all previous frames. Intuitively, each mask can be thought of as possibly a new temporal object to be tracked. 

As there are no propagated masks, we ignore the mask filtering step in stage 2 and move to frame-level scene graph generation in stage 3. Formally, given only the input frame $f_1$, Gemini produces a complete frame-level scene graph $G_1=(O, R)$. Each object $o_k \in O$ is identified by a Gemini-assigned ID, a class label (e.g. bottle, hand, table), and bounding box coordinates. Each relationship $r_q \in R$ is denoted by the subject's ID, a predicate label (e.g.\ on, in, holding), and the object's ID.

Because we are only interested in objects identified from the Gemini scene graph, we employ a matching algorithm to map each Gemini object $o_k \in O$ with a SAM2 mask $m_j \in M$. This is object-mask matching in stage 4. That is, for each $o_k$, we find an $\hat{m_k}$ satisfying

$$\hat{m_k}=\max_{m_j}\{IoU(o_k^b, m_j^b)\}$$

\noindent where $IoU$ denotes the Intersection over Union algorithm, and $o_k^b$ and $m_j^b$ are bounding boxes for object $o_k$ and mask $m_j$. To address false matches, we also threshold the $IoU$ of $o_k$ and $\hat{m_k}$ to be above 0.1. Otherwise, we discard $o_k$ and all incident relationships that contain $o_k$ as a subject or object. For example, suppose the frame depicts a small phone on a kitchen counter such that their bounding boxes have $IoU < 0.1$. If Gemini correctly grounds a phone object but SAM2 fails to segment a phone mask, we would ignore the Gemini phone object and its relationships instead of attempting to match it with SAM2's kitchen counter mask. In other words, although this matching algorithm typically removes irrelevant masks, we occasionally remove objects (and their relationships) as well to avoid bad object-mask matches. At this point, we have produced a subset of matched objects $O' \subset O$, matched masks $M' \subset M$, and relationships $R' \subset R$. 

Since SAM2 draws more accurate bounding boxes than Gemini, we transition to SAM2 masks as the basis for objects in our synthesized scene graph. To do so, we express the matching algorithm above as the surjective function

$$f:O'\rightarrow M'$$

\noindent such that each $o_k' \in O'$ has $f(o_k')=\hat{m_k}$. This allows us to define the pseudo-inverse function as

$$f^{-1}: M'\rightarrow O'$$

\noindent such that each $\hat{m_k} \in M'$ has $f^{-1}(\hat{m_k})=o_k'$, for some arbitrary $o_k'$ among all such $\{o_k': f(o_k')=\hat{m_k}\}$. Observe $f^{-1}$ is not a true inverse because $f$ is not necessarily injective. In context, this means that a SAM2 mask that is matched with multiple Gemini objects will only prioritize one of those objects.

We are now ready to perform temporally-consistent scene graph synthesis in stage 5. Since each SAM2 mask $\hat{m_k} \in M'$ is already associated with a bounding box and tID, we extend $\hat{m_k}$ to a new object $o_k^*$ by adding the class label from a matched object $f^{-1}(\hat{m_k})=o_k'$. Additionally, for each relationship $r_q \in R'$ that is associated with triplet $\langle o_s', p_q, o_t' \rangle$, we replace objects $o_s'$ and $o_t'$ with masks $f(o_s')=\hat{m_s}$ and $f(o_t')=\hat{m_t}$. As above, we extend these masks  to $o_s^*$ and $o_t^*$ in order to obtain new relationship $r_q^*=\langle o_s^*, p_q, o_t^* \rangle$. In practice, this is equivalent to replacing the Gemini-assigned object IDs with the tIDs of matching masks. We finally return the set of all such objects $o_k^*$ and relationships $r_q^*$ as the temporally-consistent scene graph in frame 1.

\subsection{Mask-Propagated Scene Graph Generation}  \label{following_frames}

For each of the following frames, we begin with mask propagation in stage 1. Denoting the current frame as $f_{i+1}$, we inductively suppose that all frames $f_1,...,f_i$ have already produced sets of matched masks $M_1',...,M_i'$. We add the tID, mask segmentation, and frame index of each matched mask $\hat{m_k} \in \bigcup_{j=1}^{i}M_j'$ to the inference state of SAM2's video predictor class. In reality, we speed this step up by only adding newly matched masks $M_j'' \subset M_j'$ at every frame. SAM2 then propagates these masks to the current frame, obtaining a set of propagated masks $P$. Observe that each propagated mask $p_l \in P$ has the same tID as some previous mask $\hat{m_l} \in \bigcup_{j=1}^{i}M_j'$.

We now move to mask generation \& filtering in stage 2. SAM2 first generates a set of masks $M:=M_{i+1}$. Recall that each $m_j \in M$ has a completely unique tID among all masks in frames $f_1, ..., f_{i+1}$. In the filtering step, we combine $P$ and $M$, then remove any generated mask $m_j \in M$ that overlaps significantly with masks in $P$. We reason that this maintains temporal consistency because SAM2 masks now include both new temporal objects (generated masks in $M$) and existing temporal objects (propagated masks in $P$). Concretely, we define the produced set of masks to be

$$M_{new}=P \cup \{m_j \in M: overlap(m_j, P_{mask}) < 0.5\}$$

\noindent where $P_{mask}$ denotes the bitwise OR of all propagated masks in $P$. Here, $overlap$ is calculated as the segmentation-based formula $\frac{\text{area of } m_j \cap P_{mask}}{\text{area of }m_j}$.

Substituting $M_{new}$ for $M$, we finish stage 2. As we did in frame 1 (see Sec. \ref{first_frame}), we continue to frame-level scene graph generation in stage 3, object-mask matching in stage 4, and temporally-consistent scene graph synthesis in stage 5 . Frame-by-frame, we repeat these 5 stages of the pipeline until all temporally-consistent scene graphs are produced.

%% file: sec/3_experiments.tex
\section{Experiments}
\label{sec:experiments}

\begin{figure*}
  \centering
   \includegraphics[width=0.95\linewidth]{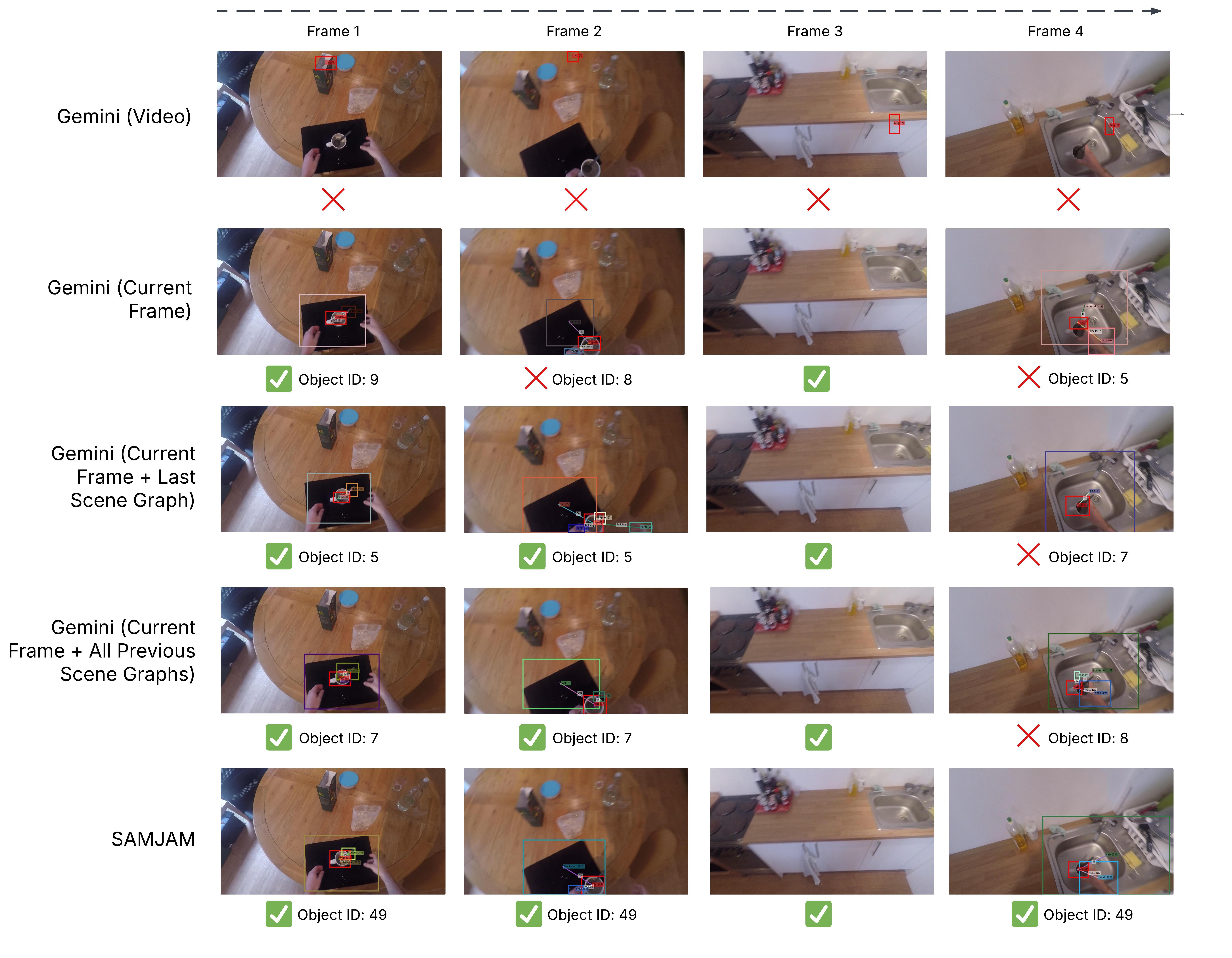}

   \caption{Qualitative results. We evaluate VidSGG models using a brief video clip taken from EPIC-KITCHENS \cite{Damen2018EPICKITCHENS} that shows a mug being moved. Illustrated above are the trimmed scene graph outputs on 4 frames from the clip, with bounding boxes for the mug highlighted in red. For Gemini (Video), object grounding of the mug completely fails. For all other methods, we display the object IDs assigned to the mug at each frame. Notably, SAMJAM is the only method that produces a consistent object ID across all 4 frames.}
   \label{fig:qualitative}
\end{figure*}

\begin{table*}
  \centering
  \begin{tabular}{@{}lcccc@{}}
    \toprule
    \textbf{Method} & \textbf{Clip 1 Recall} & \textbf{Clip 2 Recall} & \textbf{Clip 3 Recall} & \textbf{Mean Recall}\\
    \midrule
    Gemini (Current Frame) & 12.5 & 2.08 & 5.61 & 6.73 \\
    Gemini (Current Frame + Last Scene Graph) & 27.16 & 9.59 & 18.18 & 18.31 \\
    Gemini (Current Frame + All Previous Scene Graphs) & 45.33 & 10.77 & \textbf{37.88} & 31.33 \\
    SAMJAM & \textbf{53.73} & \textbf{35.42} & 29.84 & \textbf{39.66}  \\
    \bottomrule
  \end{tabular}
  \caption{Experiment results. Recall is calculated for each clip, respectively "pick up cereal bag", "throw milk carton in bin", and "cut bell pepper".}
  \label{tab:recall}
\end{table*}

In this section, we conduct both qualitative and quantitative experiments. Qualitatively, we compare how different models track the ID of a moving cup in a short, dynamic video. The models evaluated can be broadly split up into Gemini 2.0 Flash (using different input strategies) and SAMJAM. Specifically, these models are (1) Gemini using the entire video, (2) Gemini using only the current frame, (3) Gemini using the current frame plus the last scene graph, (4) Gemini using the current frame plus all previously generated scene graphs, and (5) SAMJAM (our method). Furthermore, we empirically evaluate 4 of the 5 models above on egocentric kitchen videos and report our findings.

\subsection{Dataset} We evaluate our method on 3 video clips taken from the EPIC-KITCHENS and EPIC-KITCHENS-100 datasets \cite{Damen2018EPICKITCHENS, Damen2022RESCALING}. 11 frames are sampled from each clip to capture the course of an action. From easy to hard, the actions being evaluated are pick up cereal bag, throw milk carton in bin, and cut bell pepper.

\subsection{Metric} Recall@K, which considers the the top-K confident relationship predictions, is the conventional metric for evaluating SGG \cite{LI2024127052} and VidSGG \cite{nguyen2024motion} models. We note that precision-based metrics, such as Mean Average Precision (mAP), are pessimistic since it is infeasible to exhaustively annotate all possible relationships in our ground truth \cite{10.1007/978-3-319-46448-0_51}. In our case, the absence of explicit confidence scoring in VLMs makes it difficult to select K. Thus, we calculate the standard recall, which takes into account all predicted relationship at every frame. Averaging out the recall across all videos arrives us at the mean recall.

We define a true positive prediction to satisfy 3 criteria, subject to human judgment. First, each relationship must be a correctly labelled triplet, i.e. \textlangle subject, predicate, object\textrangle. Second, each object must be contained in a reasonably tight bounding box. Third, each object must be assigned a consistent ID, with regards to the object's assigned ID at time of first correct identification. This ID must also be unique.

\subsection{Results} \label{results}
We start by reading Figure \ref{fig:qualitative}, which provides valuable qualitative insights into the performance of Gemini 2.0 Flash and SAMJAM. This will also help explain the quantitative results later. From top to bottom, the models from Figure \ref{fig:qualitative} are:

\textbf{Gemini (Video)}:  We remark that Gemini completely fails to ground the mug, which is indicative of its poor bounding box generation in general. This behaviour is possibly due to Gemini being primarily trained on video summarization when fed video inputs, with limited emphasis on frame-level object grounding. We note that because of its poor object grounding performance, we do not run experiments on Gemini (Video) in Table \ref{tab:recall}.

\textbf{Gemini (Current Frame)}: This approach assigns different IDs to the mug in all frames where the mug is present. In fact, ID instability is a common theme with the current frame input strategy: without knowledge of any other frame, Gemini has no temporal context and assigns random, meaningless IDs to objects in each scene graph.

\textbf{Gemini (Current Frame + Last Scene Graph)}: We remark that the ID of the mug is now consistent across the first 2 frames but changes in the 4th frame. While this method outperforms inputting the current frame alone, we still lose temporal context in cases where an object disappears from view then reappears some frames later (e.g. the mug's ID is forgotten in frame 4 because it goes out of view in frame 3). At other times, objects that have disappeared from view can cause previously assigned IDs to be repurposed for new objects, creating temporal inconsistency.

\textbf{Gemini (Current Frame + All Previous Scene Graphs)}: We expect this method to perform the best since it is provided full temporal context. However, we observe that Gemini still fails to track the mug's ID in frame 4. We reason that, since Gemini usually generates integer IDs from 1 to $|O|$ (size of object set) in each frame, the model has trouble reconciling these new IDs with the old scene graph IDs. Similar to Gemini (current frame + last scene graph), this also leads to old IDs being repurposed for new objects. Moreover, we note that inputting many previous scene graphs degrades overall object detection compared to inputting the current frame alone, as Gemini sometimes hallucinates by tracking objects that have already disappeared from view.

\textbf{SAMJAM}: Our proposed method. Not only do we consistently track the mug's ID across all frames, we also produce the tightest bounding box for the mug at each frame compared to every other approach.

We briefly mention our experience with other VidSGG methods that were not shown in Figure \ref{fig:qualitative}. For instance, inputting multiple images into Gemini causes object grounding to fail completely. Additionally, we find that VideoLLaMA3 \cite{zhang2025videollama3frontiermultimodal} -- an open-source SOTA video-understanding model -- cannot generate an interpretable scene graph for our purpose.

We are now ready to interpret the quantitative results. Table \ref{tab:recall} reports recall on three clips -- “pick up cereal bag,” “throw milk carton in bin,” and “cut bell pepper” -- as well as the mean recall across all clips. Recall improves for Gemini when we provide more context in our input, with the method that leverages all previous scene graphs yielding markedly better performance than other Gemini configurations. Nevertheless, SAMJAM attains the highest recall on two of the three tasks and achieves the best overall mean recall (39.66\%), outperforming the next-best method by over 8\%. 

\subsection{Limitations} SAMJAM is limited by Gemini's grounding accuracy. While generating qualitative results in Figure \ref{fig:qualitative}, we noticed instances where inaccurate bounding boxes produced by Gemini led SAMJAM to match the mug with another object instead (e.g. the spoon). In other words, while this method rectifies small bounding box errors from Gemini, large errors remain unaddressed and could result in false matches for the synthesized scene graph. Another source of error comes from mistakes in Gemini's zero-shot object classification. For example, the transparent cereal bag is sometimes labelled as glass in our quantitative experiment.

With regards to efficiency, we note that SAMJAM is bottlenecked by SAM2's automatic mask generation in stage 2 (which takes over 20 seconds per frame). Although mask generation parameters can be fine-tuned to accelerate this process, we leave a comprehensive analysis of the performance-efficiency trade offs for future work.

%% file: sec/4_conclusion.tex
\section{Conclusion}
\label{sec:conclusion}

In this work, we addressed the challenge of generating accurate, temporally-consistent scene graphs in dynamic kitchen videos. While models such as Gemini 2.0 Flash excel at zero-shot scene understanding, they struggle with maintaining stable object identities across frames and producing tight bounding boxes. To overcome this limitations, we introduced SAMJAM, a novel pipeline that combines Gemini’s open-vocabulary detection with SAM2’s robust video-level mask propagation and frame-level segmentation. By matching Gemini-detected objects with SAM2 masks, SAMJAM ensures stable object tracking and refined spatial grounding -- ultimately producing coherent scene graphs throughout the video. Notably, our approach achieves a 8.33\% improvement in mean recall compared to using Gemini alone. In the future, we plan to evaluate our method more thoroughly on larger datasets, such as the Panoptic Video Scene Graph Generation dataset \cite{Yang_Peng_Li_Guo}, and to extend its application to improve question-answering models for egocentric food videos.

%% file: main.bbl
\begin{thebibliography}{21}
\providecommand{\natexlab}[1]{#1}
\providecommand{\url}[1]{\texttt{#1}}
\expandafter\ifx\csname urlstyle\endcsname\relax
  \providecommand{\doi}[1]{doi: #1}\else
  \providecommand{\doi}{doi: \begingroup \urlstyle{rm}\Url}\fi

\bibitem[Damen et~al.(2018)Damen, Doughty, Farinella, Fidler, Furnari, Kazakos, Moltisanti, Munro, Perrett, Price, and Wray]{Damen2018EPICKITCHENS}
Dima Damen, Hazel Doughty, Giovanni~Maria Farinella, Sanja Fidler, Antonino Furnari, Evangelos Kazakos, Davide Moltisanti, Jonathan Munro, Toby Perrett, Will Price, and Michael Wray.
\newblock Scaling egocentric vision: The epic-kitchens dataset.
\newblock In \emph{Proceedings of the European Conference on Computer Vision (ECCV)}, pages 753--771, 2018.

\bibitem[Damen et~al.(2022)Damen, Doughty, Farinella, Furnari, Ma, Kazakos, Moltisanti, Munro, Perrett, Price, and Wray]{Damen2022RESCALING}
Dima Damen, Hazel Doughty, Giovanni~Maria Farinella, Antonino Furnari, Jian Ma, Evangelos Kazakos, Davide Moltisanti, Jonathan Munro, Toby Perrett, Will Price, and Michael Wray.
\newblock Rescaling egocentric vision: Collection, pipeline and challenges for epic-kitchens-100.
\newblock \emph{International Journal of Computer Vision (IJCV)}, 130:\penalty0 33–55, 2022.

\bibitem[Google(2025)]{google2025gemini}
Google.
\newblock Gemini (gemini 2.0 flash) [large language model].
\newblock \url{https://ai.google.dev/gemini-api/docs/models/gemini}, 2025.
\newblock Accessed: March 14, 2025.

\bibitem[Hori et~al.(2023)Hori, Peng, Harwath, Liu, Ota, Jain, Corcodel, Jha, Romeres, and {Le Roux}]{Hori2023StyletransferBS}
Chiori Hori, Puyuang Peng, David Harwath, Xinyu Liu, Kei Ota, Siddarth Jain, Radu Corcodel, Devesh~K. Jha, Diego Romeres, and Jonathan {Le Roux}.
\newblock {Style-transfer based Speech and Audio-visual Scene understanding for Robot Action Sequence Acquisition from Videos}.
\newblock In \emph{Proceedings of Interspeech}, pages 4663--4667, 2023.

\bibitem[Ji et~al.(2020)Ji, Krishna, Fei-Fei, and Niebles]{Ji_2020_CVPR}
Jingwei Ji, Ranjay Krishna, Li Fei-Fei, and Juan~Carlos Niebles.
\newblock Action genome: Actions as compositions of spatio-temporal scene graphs.
\newblock In \emph{Proceedings of the IEEE/CVF Conference on Computer Vision and Pattern Recognition (CVPR)}, pages 10233--10244. IEEE, 2020.

\bibitem[Kim et~al.(2025)Kim, Yoon, In, Jeon, Moon, Kim, and Park]{kim2025weakly}
Kibum Kim, Kanghoon Yoon, Yeonjun In, Jaehyeong Jeon, Jinyoung Moon, Donghyun Kim, and Chanyoung Park.
\newblock Weakly supervised video scene graph generation via natural language supervision.
\newblock In \emph{Proceedings of the Thirteenth International Conference on Learning Representations}, 2025.

\bibitem[Li et~al.(2024)Li, Zhu, Zhang, Jiang, Dang, Hou, Shen, Zhao, Shah, and Bennamoun]{LI2024127052}
Hongsheng Li, Guangming Zhu, Liang Zhang, Youliang Jiang, Yixuan Dang, Haoran Hou, Peiyi Shen, Xia Zhao, Syed Afaq~Ali Shah, and Mohammed Bennamoun.
\newblock Scene graph generation: A comprehensive survey.
\newblock \emph{Neurocomputing}, 566:\penalty0 127052, 2024.

\bibitem[Lu et~al.(2016)Lu, Krishna, Bernstein, and Fei-Fei]{10.1007/978-3-319-46448-0_51}
Cewu Lu, Ranjay Krishna, Michael Bernstein, and Li Fei-Fei.
\newblock Visual relationship detection with language priors.
\newblock In \emph{Proceedings of the European Conference on Computer Vision (ECCV)}, pages 852--869. Springer, 2016.

\bibitem[Mangalam et~al.(2023)Mangalam, Akshulakov, and Malik]{mangalam2023egoschema}
Karttikeya Mangalam, Raiymbek Akshulakov, and Jitendra Malik.
\newblock Egoschema: A diagnostic benchmark for very long-form video language understanding.
\newblock In \emph{Advances in Neural Information Processing Systems}, pages 46212--46244. Curran Associates, Inc., 2023.

\bibitem[Nagar et~al.(2024)Nagar, Jaiswal, and Tan]{Nagar_Jaiswal_Tan_2024}
Aishik Nagar, Shantanu Jaiswal, and Cheston Tan.
\newblock Zero-shot visual reasoning by vision-language models: Benchmarking and analysis.
\newblock In \emph{Proceedings of the International Joint Conference on Neural Networks (IJCNN)}, pages 1--8, 2024.

\bibitem[Nguyen et~al.(2024)Nguyen, Nguyen, Li, Cothren, Yilmaz, and Luu]{nguyen2024cyclo}
Trong-Thuan Nguyen, Pha Nguyen, Xin Li, Jackson Cothren, Alper Yilmaz, and Khoa Luu.
\newblock Cyclo: Cyclic graph transformer approach to multi-object relationship modeling in aerial videos.
\newblock In \emph{Advances in Neural Information Processing Systems}, pages 90355--90383. Curran Associates, Inc., 2024.

\bibitem[Nguyen et~al.(2025)Nguyen, Wu, Bin, Nguyen, Ng, and Luu]{nguyen2024motion}
Thong~Thanh Nguyen, Xiaobao Wu, Yi Bin, Cong-Duy~T Nguyen, See-Kiong Ng, and Anh~Tuan Luu.
\newblock Motion-aware contrastive learning for temporal panoptic scene graph generation.
\newblock In \emph{Proceedings of the AAAI Conference on Artificial Intelligence}, 2025.

\bibitem[Pu et~al.(2024)Pu, Chen, Wu, Lu, and Lin]{10375886}
Tao Pu, Tianshui Chen, Hefeng Wu, Yongyi Lu, and Liang Lin.
\newblock Spatial–temporal knowledge-embedded transformer for video scene graph generation.
\newblock \emph{IEEE Transactions on Image Processing}, 33:\penalty0 556--568, 2024.

\bibitem[Ravi et~al.(2025)Ravi, Gabeur, Hu, Hu, Ryali, Ma, Khedr, R{\"a}dle, Rolland, Gustafson, Mintun, Pan, Alwala, Carion, Wu, Girshick, Dollar, and Feichtenhofer]{ravi2025sam}
Nikhila Ravi, Valentin Gabeur, Yuan-Ting Hu, Ronghang Hu, Chaitanya Ryali, Tengyu Ma, Haitham Khedr, Roman R{\"a}dle, Chloe Rolland, Laura Gustafson, Eric Mintun, Junting Pan, Kalyan~Vasudev Alwala, Nicolas Carion, Chao-Yuan Wu, Ross Girshick, Piotr Dollar, and Christoph Feichtenhofer.
\newblock {SAM} 2: Segment anything in images and videos.
\newblock In \emph{Proceedings of the Thirteenth International Conference on Learning Representations}, 2025.

\bibitem[Shang et~al.(2017)Shang, Ren, Guo, Zhang, and Chua]{Shang_Ren_Guo_Zhang_Chua_2017}
Xindi Shang, Tongwei Ren, Jingfan Guo, Hanwang Zhang, and Tat-Seng Chua.
\newblock Video visual relation detection.
\newblock In \emph{Proceedings of the 25th ACM International Conference on Multimedia}, pages 1300--1308, New York, NY, USA, 2017. Association for Computing Machinery.

\bibitem[Tang et~al.(2024)Tang, Bi, Xu, Song, Liang, Wang, Zhang, An, Lin, Zhu, Vosoughi, Huang, Zhang, Liu, Feng, Zheng, Zhang, Luo, Luo, and Xu]{vlm_survery}
Yunlong Tang, Jing Bi, Siting Xu, Luchuan Song, Susan Liang, Teng Wang, Daoan Zhang, Jie An, Jingyang Lin, Rongyi Zhu, Ali Vosoughi, Chao Huang, Zeliang Zhang, Pinxin Liu, Mingqian Feng, Feng Zheng, Jianguo Zhang, Ping Luo, Jiebo Luo, and Chenliang Xu.
\newblock Video understanding with large language models: A survey, 2024.

\bibitem[Wu et~al.(2024)Wu, Li, Chen, and Li]{wu2024longvideobench}
Haoning Wu, Dongxu Li, Bei Chen, and Junnan Li.
\newblock Longvideobench: A benchmark for long-context interleaved video-language understanding.
\newblock In \emph{Advances in Neural Information Processing Systems}, pages 28828--28857. Curran Associates, Inc., 2024.

\bibitem[Yamakata et~al.(2022)Yamakata, Ishino, Sunto, Amano, and Aizawa]{skeleton_recipe}
Yoko Yamakata, Akihisa Ishino, Akiko Sunto, Sosuke Amano, and Kiyoharu Aizawa.
\newblock Recipe-oriented food logging for nutritional management.
\newblock In \emph{Proceedings of the 30th ACM International Conference on Multimedia}, pages 6898--6904, New York, NY, USA, 2022. Association for Computing Machinery.

\bibitem[Yang et~al.(2018)Yang, Lu, Lee, Batra, and Parikh]{yang2018graph}
Jianwei Yang, Jiasen Lu, Stefan Lee, Dhruv Batra, and Devi Parikh.
\newblock Graph r-cnn for scene graph generation.
\newblock In \emph{Proceedings of the European Conference on Computer Vision (ECCV)}, pages 670--685, 2018.

\bibitem[Yang et~al.(2023)Yang, Peng, Li, Guo, Chen, Li, Ma, Zhou, Zhang, Loy, and Liu]{Yang_Peng_Li_Guo}
Jingkang Yang, Wenxuan Peng, Xiangtai Li, Zujin Guo, Liangyu Chen, Bo Li, Zheng Ma, Kaiyang Zhou, Wayne Zhang, Chen~Change Loy, and Ziwei Liu.
\newblock Panoptic video scene graph generation.
\newblock In \emph{Proceedings of the IEEE/CVF Conference on Computer Vision and Pattern Recognition (CVPR)}, pages 18675--18685, Vancouver, BC, Canada, 2023. IEEE.

\bibitem[Zhang et~al.(2025)Zhang, Li, Cheng, Hu, Yuan, Chen, Leng, Jiang, Zhang, Li, Jin, Zhang, Wang, Bing, and Zhao]{zhang2025videollama3frontiermultimodal}
Boqiang Zhang, Kehan Li, Zesen Cheng, Zhiqiang Hu, Yuqian Yuan, Guanzheng Chen, Sicong Leng, Yuming Jiang, Hang Zhang, Xin Li, Peng Jin, Wenqi Zhang, Fan Wang, Lidong Bing, and Deli Zhao.
\newblock Videollama 3: Frontier multimodal foundation models for image and video understanding, 2025.

\end{thebibliography}
